# Deep learning model for Mongolian Citizens' Feedback Analysis using Word Vector Embeddings

Zolzaya Dashdorj [12]*, Tsetsentsengel Munkhbayar [1] and Stanislav Grigorev [2]

1   Mongolian University of Science and Technology; zolzaya@must.edu.mn, tsetsee.yugi@gmail.com

2   Irkutsk National Research Technical University; svg@istu.edu

*Correspondence: zolzaya@must.edu.mn;

**Abstract**

A large amount of feedback was collected over the years. Many feedback analysis models have been developed focusing on the English language. Recognizing the concept of feedback is challenging and crucial in languages which do not have applicable corpus and tools employed in Natural Language Processing (i.e., vocabulary corpus, sentence structure rules, etc). However, in this paper, we study a feedback classification in Mongolian language using two different word embeddings for deep learning. We compare the results of proposed approaches. We use feedback data in Cyrillic collected from 2012-2018. The result indicates that word embeddings using their own dataset improve the deep learning based proposed model with the best accuracy of 80.1% and 82.7% for two classification tasks.

## 1. Introduction

Increasing number of user opinions about products or services shared throughout web sources and social networks has attracted many researchers to conduct various research studies in the field of sentiment analysis [1-3, 9-14] and classification tasks [4-8] by determining user attitudes on social media, categorizing emails, automatically marking customer questions, and categorizing news articles. Most deep learning methods were more effective in Natural Language Processing. However, most techniques are demonstrated in the English language. In other languages, a less vocabulary corpus is developed and the Natural Language Processing tools are insufficient. Therefore, the amount of labeled data is insufficient and requires more comprehensive approaches to use the labeled data in available languages for solving the sentiment classification problems in the other languages. These challenges have led to an interesting research direction in sentiment analysis using word vector embeddings from the cross-language perspectives [12, 13] that has become an emerging issue in the sentiment analysis community [14]. FastText word embedding model is well used in sentiment analysis and feedback classification tasks [8,15]. A single layered deep learning based two classification models are employed using FastText word embeddings [15]. Compared to other models with complex architectures, the single layered models provide competitive results. The best accuracy is produced by the fastText combined with the Convolutional Neural Network model, with 80-84% accuracy for two different feedback dataset. This study also indicates that the use of fastText embedding can improve the performance of the single-layered BiLSTM model. Our research aims at analyzing Mongolian feedback that is written in Cyrillic and performing deep learning based classification tasks using word embeddings through neural networks. The feedback has been collected at the Mongolian government's citizens and public relation centre. The Mongolian government's citizens and public relation centre has introduced an information system that receives citizens' feedback about requests, complaints, criticisms and appreciation, distributes them to relevant government agencies, and monitors decisions. Citizens' feedback is received through more than ten channels, including calls, websites, in-person, text messages, Facebook, mobile, Twitter, email, kiosks, and skype. However, operators struggle in understanding the large amount of feedback and distributing them to the relevant agencies daily. Therefore, we demonstrate two classification tasks using Convolutional Neural Networks and Bidirectional LSTM that performed relatively well in other studies in English feedback. This type of analysis encourages customer relation management applications in other languages.

2. **Data Collection**

We conducted a scraped dataset between October 2012 and November 2018 at the Government Citizens' Public Relations Center System. A total of 78,810 feedback were received and resolved by 76 relevant government agencies. Figure 1 shows the percentage of feedback types. 70% of the feedback is related to the feedback type.

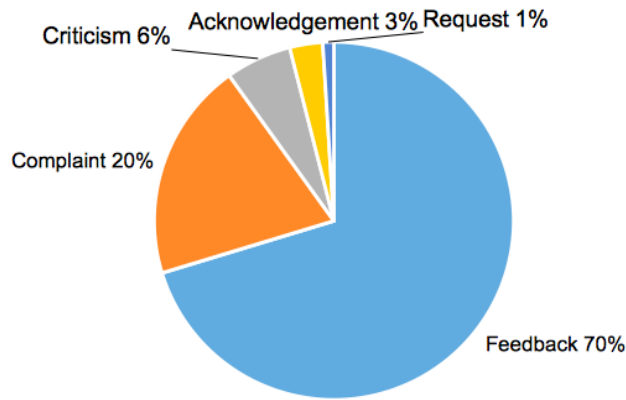

Figure 1. Percentage of citizens' feedback types

The feedback data is divided into five main categories: comment, complaint, criticism, request, and compliment, including the textual content of the feedback. Figure 2 presents the distribution of the feedback over government agencies. Most feedback corresponds to the City Governor's Office.

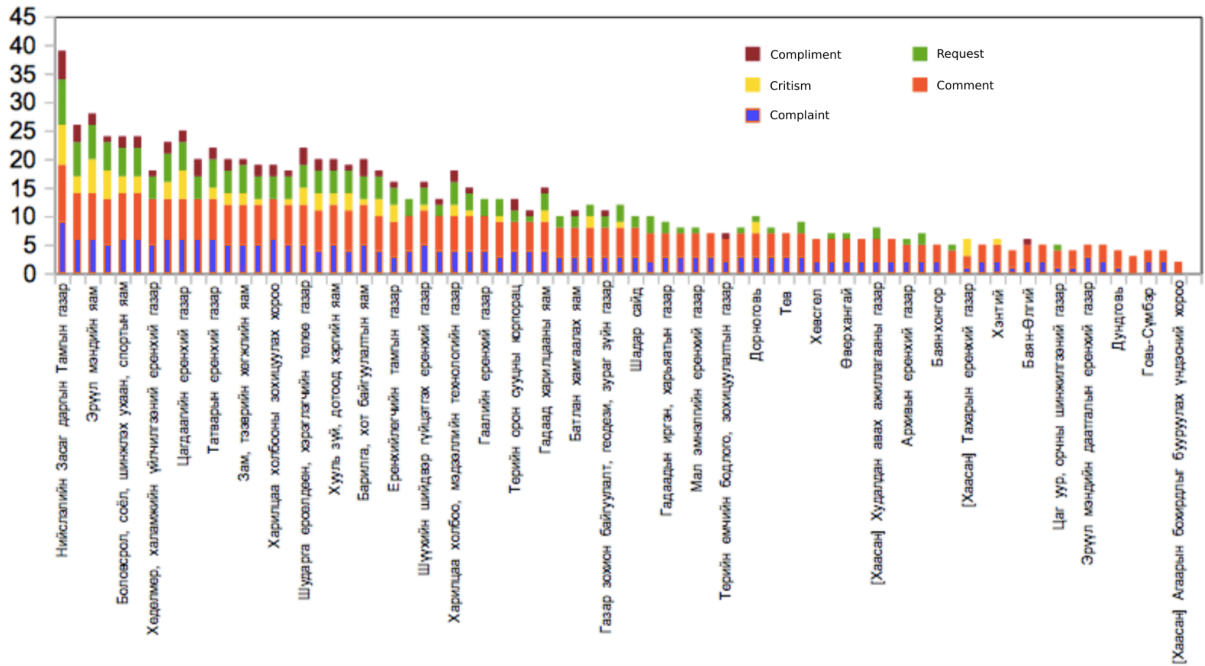

Figure 2. Citizens' feedback distribution over government agencies

### 3. Methodology

Citizens' feedback is given in the free-form text. We use natural language processing techniques to understand the content type of feedback. We clean the data, remove unnecessary words and apply word embedding vector models. A vector model converts text into numeric vectors, allowing mathematical operations to be performed. Once the feedback are vectorized, we apply deep learning techniques, such as Convolutional Neural Network and Bidirectional Long Short-Term Memory.

**Word Vector Embedding Model.** We summarize the word vector models used in the embedding layer of the neural networks models.
- FastText vector model [8]. The vector model of words taught in Mongolian text on Wikipedia and Common Crawl, is 300-dimensional and contains 600,000 Mongolian words. The model was trained using CBOW with position weights with character n-grams of length 5, a window of size 5 and 10 negatives.
- Feedback Word2Vec word vector model. We created our word vector model with a corpus of 71,177 feedback, subtracting Latin content from the total dataset. We created a 300-dimensional model with a vocabulary of 86,956 words.

**Neural Networks models.** We employ two neural networks models for learning feedback data with an embedding layer. We summarize the architectural design of the network models.

- Convolutional Neural Network (CNN). CNN is a deep learning method and consists of layers. We experimented with combining a CNN model with a word embedding model. We configured the architecture of CNN as a multi-layer network. Sequentially eight layers were designed with the following hyper-parameters. The first layer is an embedding layer with the 300 dimension of 255 length words, and the second layer is a 1 dimensional convolutional layer with the 300 filters, 5 kernels, Relu activation, and the 3rd layer is a one dimensional convolutional layer with the 300 units, 4 kernels, Relu activation, and 4th layer is a global max pooling layer, and 5th layer is a dense layer with 300 units, Relu activation, and 6th layer is a dropout layer with the dropout rate of 0.5, and 7th layer is a dropout layer with the dropout rate of 0.5, and an output layer contains 'softmax' activation and the number of output classes, and an output layer contains 'softmax' activation and the number of output classes.
- Bidirectional Long Short-Term Memory (BiLSTM) is a sequence processing model that consists of two LSTMs with the sequence of directions two-way; one taking the input in a forward direction, and the other in a backwards direction. We configured the architecture of BiLSTM as a multi-layer network. Sequentially seven layers were designed with the following hyper-parameters. The first layer is an embedding layer with the 300 dimension of 255 length words, and the second layer is a 1 dimensional spatial dropout layer with the dropout of 0.2, and the 3rd layer is a bidirectional LSTM layer with the 300 units of forward direction, 4th layer is a bidirectional LSTM layer (backward direction) with the 150 units, dropout rate of 0.2, recurrent dropout of 0.2, and 5th layer is a dense layer with the 150 units, activation function as Relu, and 6th layer is a dropout layer with a dropout rate of 0.5, an output layer contains 'softmax' activation and the number of output classes.

4. **Experiment results**

We perform two classification tasks; one is a prediction of government agencies giving citizens feedback. The second is a prediction of emotional types given citizens' feedback. The distribution of feedback to government agencies is arbitrary and imbalanced. So we have chosen agencies that gathered between 1,000 and 2,000 feedback. Therefore, 12 agencies are selected with a total of 15,530 feedback. However, most feedback was written in Cyrillic, but some were written in Latin letters. So we cleaned the Latin feedback, and the dataset was reduced to 13,744 feedback. The feedback was tokenized, removing a stop word that did not significantly affect the result. There are 75 useless words in this list, such as 'I', 'you', 'you', 'they', 'we', 'candle', 'candle', 'he', 'this', and so on. The tokens are unified, all letters are minimized, and spaces are deleted. Noise Removal or unnecessary characters removed. For example, '!', '?', '"', '.', ',', '-',… and so on. Also, words that are misspelled and do not show any meaning are filtered out. 677 tokens such as 'color', 'nubyn', 'sbd', 'uune', 'zgn', 'not', 'zg', 'shzg', 'ukhaayas', and 'deune' were removed from the training data. To perform the classification task, the dataset is divided into training -70%, testing - 20% and validation - 10%. Table 1 summarizes the classification performance results.

Table 1. The government agency prediction performance

| Algorithm | Epoch (Early Stop) | Precision | Recall | F1-Micro | Accuracy |
|---|---|---|---|---|---|
| CNN + FastText | 34 | 71.7% | 71.7% | 71.7% | 73.4% |
| CNN + Feedback Word2Vec | 35 | 72.0% | 72.0% | 72.0% | 72.0% |
| BiLSTM+ FastText | 34 | 76.1% | 76.1% | 76.1% | 76.1% |
| BiLSTM+ Feedback Word2Vec | 23 | 80.1% | 80.2% | 80.7% | 80.7% |

The BiLSTM model with an embedding of Feedback Word2Vec performed well with an accuracy of 80.7% compared to the CNN model. The result indicates that feedback contents can be predicted enough to replace human jobs at the citizen relation center. Another role of the human job at the citizen relation center is assigning emotional feedback. Because the distribution of the feedback to government agencies is arbitrary and imbalanced in terms of emotional feedback, we combine comment types into two emotional categories. A total of five types of feedback were aggregated into two categories. Complaints and suggestions were combined to create a neutral category, and criticisms and complaints were combined to form a negative category. The gratitude was excluded from the experiment due to insufficient classification data. In order to make the test results more reliable, 10,000 data and no more than 500 words of feedback were selected from each category to maintain balance. The result is a total of 20,000 data in the two categories. As presented in Table 2, the model was performed and validated in a data split of 80/20 with a k-fold (k=5) validation.

Table 2. The comment type prediction performance result

| Algorithm | Epoch (Early Stop) | Precision | Recall | F1-Micro | Accuracy |
|---|---|---|---|---|---|
| CNN + FastText | 22 | 80.4% | 80.4% | 80.4% | 80.4% |
| CNN + Word2Vec | 22 | 79.7% | 79.7% | 79.7% | 79.7% |
| BiLSTM+ FastText | 21 | 80.9% | 80.9% | 80.9% | 80.9% |

| BiLSTM+ Word2Vec | 22 | 82.1% | 82.1% | 82.1% | 82.1% |

The experiment result indicated that, like our previous experiment, the BiLSTM + Word2Vec model performed a successful result of 82.1% accuracy.

5. **Conclusion**

The Government of Mongolia receives comments, criticism, compliments, complaints and requests from citizens. It distributes them to the appropriate authorities, slowing down the process of responding to errors made by operators and citizens. Therefore, this process has been made more intelligent by reducing machine involvement using machine learning methods. One solution is to create a system that automatically categorizes feedback content. The experiments used deep learning methods (CNN, BiLSTM) in combination with word embeddings (FastText, Feedback Word2Vec). From this, the Feedback Word2Vec model, based on the vocabulary generated from the training data, performed better with the BiLSTM architecture than the deep learning models with the FastText model. The BiLSTM + Feedback Word2Vec model performed an 80.7% and 82.1% accuracy for government agency prediction given feedback and emotion type prediction, respectively. We further improve the accuracy of the models using other advanced techniques in deep learning.

**Acknowledgement**

This research was supported by Mongolian Foundation for Science and Technology (MFST), project number "STIPICD-2021/475". The study was supported by the Mongolian Ministry of Education and Science and also partly with a grant from Irkutsk National Research Technical University.